\documentclass[pmlr]{jmlr}% new name PMLR (Proceedings of Machine Learning)

\usepackage{longtable}% for long tables

\usepackage{booktabs}
\usepackage[load-configurations=version-1]{siunitx} % newer version

\makeatletter
\def\set@curr@file#1{\def\@curr@file{#1}} %temp workaround for 2019 latex release
\makeatother

\theorembodyfont{\upshape}
\theoremheaderfont{\scshape}
\theorempostheader{:}
\theoremsep{\newline}

\jmlrvolume{149}
\jmlryear{2021}
\jmlrworkshop{Machine Learning for Healthcare}

\usepackage{amsmath}
\usepackage{amssymb}
\usepackage{tikz}
\usetikzlibrary{positioning}
\usetikzlibrary{calc}
\usepackage{longtable}
\usepackage{booktabs}
\usepackage{dirtytalk}

\title[Pushing the Limits of Medical Codes Prediction from Clinical Notes by Machines]{Read, Attend, and Code: Pushing the Limits of \\ Medical Codes Prediction from Clinical Notes by Machines}

\author{\Name{Byung-Hak Kim}
      \Email{hak.kim@akasa.com}\\ 
      \Name{Varun Ganapathi}
      \Email{varun@akasa.com}\\ 
      \addr AKASA\\
      South San Francisco, CA, USA} 

\editor{Editor's name}

\begin{document}

\maketitle

\begin{abstract}
Prediction of medical codes from clinical notes is both a practical and essential need for every healthcare delivery organization within current medical systems. Automating annotation will save significant time and excessive effort spent by human coders today. However, the biggest challenge is directly identifying appropriate medical codes out of several thousands of high-dimensional codes from unstructured free-text clinical notes. In the past three years, with Convolutional Neural Networks (CNN) and Long Short-Term Memory (LTSM) networks, there have been vast improvements in tackling the most challenging benchmark of the MIMIC-III-full-label inpatient clinical notes dataset. This progress raises the fundamental question of how far automated machine learning (ML) systems are from human coders' working performance. We assessed the baseline of human coders' performance on the same subsampled testing set. We also present our Read, Attend, and Code (RAC) model for learning the medical code assignment mappings. By connecting convolved embeddings with self-attention and code-title guided attention modules, combined with sentence permutation-based data augmentations and stochastic weight averaging training, RAC establishes a new state of the art (SOTA), considerably outperforming the current best Macro-F1 by 18.7\%, and reaches past the human-level coding baseline. This new milestone marks a meaningful step toward fully autonomous medical coding (AMC) in machines reaching parity with human coders' performance in medical code prediction. 
\end{abstract}

\section{Introduction}
\label{sec:introduction}
Automatic clinical coding (ACC) and emerging AMC are technologies that use natural language processing (NLP) to generate diagnosis and procedure medical codes from clinical notes automatically. A human coder or health care provider scans the medical documentation in electronic health records (EHR), identifying essential information and annotating codes for that particular treatment or service with the current ACC engine. With a wide range of medical services and providers (primary care clinics, specialty clinics, emergency departments, mother-baby units, outpatient and inpatient units, etc.), the complexity of human coders' tasks increases as the medical industry advances, while productivity standards decrease as charts take more time to review. A typical performance for an inpatient coder is 2.5 charts per hour because the review includes capturing historical diagnoses, lab results, radiology results, and several notes from different providers seen during his or her hospital stay.

\subsection*{Generalizable Insights about Machine Learning in the Context of Healthcare}
To better understand human coders' coding performance baseline, we have hired two professional coders to code the same MIMIC-III inpatient clinical notes testing set that the ML systems were also tested on. We found that the human coders' inter-agreement rates are not as high as believed, and the RAC model-based ML system exceeds the human coding baseline by a large margin. As far as we know, this is the first paper that benchmarks human coders' performance on the MIMIC-III dataset by estimating human coders' ability to agree with one another. Our work makes the three main contributions below:

\begin{itemize}
    \item First, the human coding baseline is estimated via a primitive internal website. Given the reference diagnosis and procedure codes that Beth Israel Deaconess Medical Center (BIDMC) assigned, we have two coders independently code the different set of notes and evaluate where their total annotations have differed from the references. 
    
    \item Second, a new RAC architecture that can process unstructured medical notes and attend to text areas annotating medical codes is developed. The main building blocks of RAC are built upon self-attention and code-title guided attention modules that work on sets of sentence vectors. In principle, we look at the medical codes prediction problem as a set-to-set assignment learning problem from the set of input sentence vectors to the set of code labels and employ the problem's unique permutation equivariant property\footnote{Consider permuting the notes' sentence vectors; medical code outputs will be the same but permuted.} in the design. 
    
    \item Third, based on the MIMIC-III dataset, we demonstrate the RAC model's effectiveness in the most challenging full codes prediction testing set from \emph{inpatient} clinical notes. The RAC model wins over all the previously reported SOTA results considerably. Compared to models that integrate more priors like CNNs \& LSTMs, self-attention-based models require more data to generalize well. Hence, to train with the same sized dataset, we utilize the sentence permutation-based data augmentation and stochastic weighted average training~\citep{Izmailov18}.
    
\end{itemize}

\section{Related Work}
\label{sec:related works}
\textbf{Automatic Clinical Coding:} 
Automatic coding of clinical notes can be treated as a multilabel classification problem. The clinical coding problem has received lots of attention in the last few years from this perspective, and it was demonstrated that CNN and LSTM based deep learning models perform better than conventional machine learning models \citep{Baumel17,Shi17,Mullenbach18,Li19,Huang19,Chen19,Xie19,Li20,Vu20}; lately, the works have expanded to non-English clinical notes \citep{Azam19,Wang20,Reys20}. However, each approach has its disadvantages. The CNN based model requires a stack of many CNN layers to \say{see} the whole input, and while the LSTM based method is useful at overcoming this, it is not easily parallelizable in training and inference. 

\textbf{Clinical Transformers:} 
Converting the free-text in clinical notes into a representation that can easily be used remains one of the prime NLP challenges. With the boom of the Transformer based models, specialized domain versions are trained from scratch or adapted to domains and tasks (see \citealp{Lee20,Peng19,Alsentzer19,Gu20,Zhang20}). However, successfully applying Transformer models to the medical codes prediction problem remains an open challenge as reported in \citep{Li20,Ji20}, primarily due to three reasons: first, free-text clinical notes are unstructured, riddled with spelling errors, and consist of language particular to the medical domain. Second, the output codes space is large, and there is a severe long-tail sparsity issue with over 68,000 codes in the new ICD-10-CM system~\citep{CDC15}, for example. Third, the standard Transformer-based models' potential disadvantage in handling longer length of text input than CNN and LSTM based models, which presumably can handle texts with an arbitrary length.

\section{Human-Level Coding Baselines} 
\label{sec:Human-Level Coding Baseline}

\begin{figure*}[ht]
    \centering
    \includegraphics[width = 1.0\textwidth]{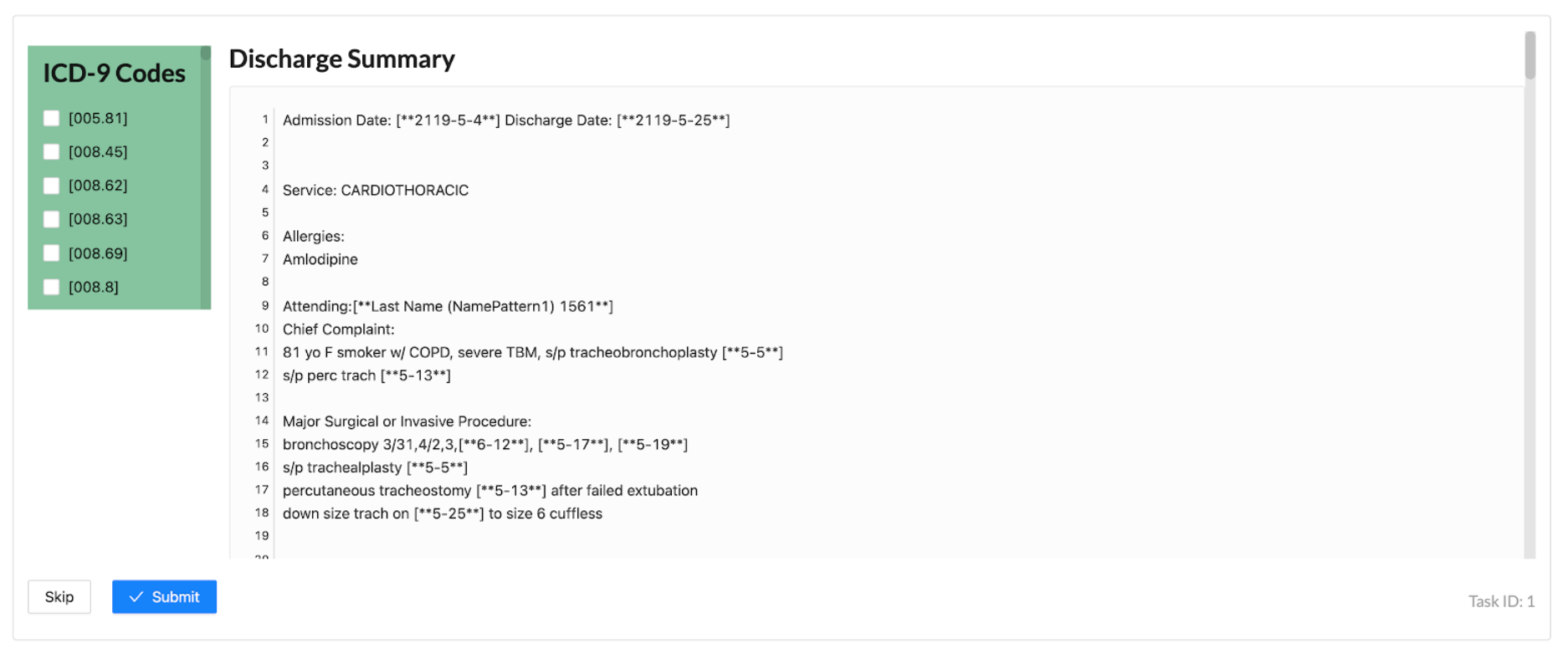}
    \caption{Our web interface for collecting human coders' assignment of a set of ICD-9 codes to inpatient discharge summaries, described in Section~\ref{sec:Human-Level Coding Baseline}. The left column lists the set of ICD-9 codes, and each discharge summary is displayed in the right panel. Note that an entire set of 4,075 codes are too long to show in the screenshot captured. There is a scroll bar next to the ICD-9 codes column that the coder can scroll to click on the appropriate code. Alternatively, the coders can use the find shortcut (e.g., Ctrl + F) and type the code to save time from scrolling to click on the appropriate code. All discharge summaries are sampled from open-access, de-identified MIMIC-III patients' data treated in intensive care and is not subject to the HIPAA Privacy Rule restrictions on sharing protected health information~\citep{Mimiciii16}.}
    \label{fig:Web}
\end{figure*}
\label{sec:Webpage screenshot}

\textbf{Evaluation Design:} 
To assess the human coders coding baselines, we hired two professional CPC certified coders with more than five years of hospital coding, including ICD-9 coding experience. We asked them to assign a set of ICD-9 codes (from a shortlisted 4,075 codes) to a total of 508 inpatient discharge summaries randomly sub-sampled from the MIMIC-III testing set for about 30 hours over a week. The website was built using the Label Studio~\citep{LS20} for this task, and each coder is required to log in with their accounts to the website to get started. The coders are given the same discharge summary as was provided to our ML system and were asked to select (not to type) the best possible set of codes in the list for the displayed discharge summaries (see Figure~\ref{fig:Web} for more details). This environment makes the tasks easier than paper coding, though less convenient than a full-fledged computer-assisted coding system. The intention is to provide the same possible conditions as for the ML system. Additionally, all coders are instructed to code as normally as possible without a time limit to spend on each chart and to use whatever resources they usually use while adhering to coding guidelines. They were asked to code continuously with as few interruptions and distractions as possible, not to communicate with other coders, and not to skip notes. 

\begin{table*}
    \caption{Medical code annotation results (in \%) by human coders compared to our RAC model-based ML system's prediction results on 508 random subsamples from the MIMIC-III-full-label testing set. Note that we evaluate the concordance of code assignments between the annotations and the human references in the testing set, as described in Section~\ref{sec:Human-Level Coding Baseline}, and the ML system achieves 3.9 times better Micro-Jaccard similarity than human coders.}
    \label{table: human perf summary}
    \centering
    \begin{tabular}[t]{l|l|l|l}
    \toprule[\heavyrulewidth]
    & \textbf{Jaccard Similarity} & \textbf{Precision} & \textbf{Recall} \\
    & Macro \ \ \ Micro & Macro \ \ \ Micro & Macro \ \ \ Micro \\
    \midrule
    Human Coders & 1.8 \ \ \ \ \ \ \ \ \ \textbf{10.7} & 3.5 \ \ \ \ \ \ \ \ \ 47.7 & \ \ \ 2.1 \ \ \ \ \ \ 12.1\\ 
    RAC Model & 6.4 \ \ \ \ \ \ \ \ \ \textbf{42.0} & 7.1 \ \ \ \ \ \ \ \ \ 63.7 & \ 10.8 \ \ \ \ \ \ 55.2\\ 
    \bottomrule[\heavyrulewidth]
    \end{tabular}
\end{table*}

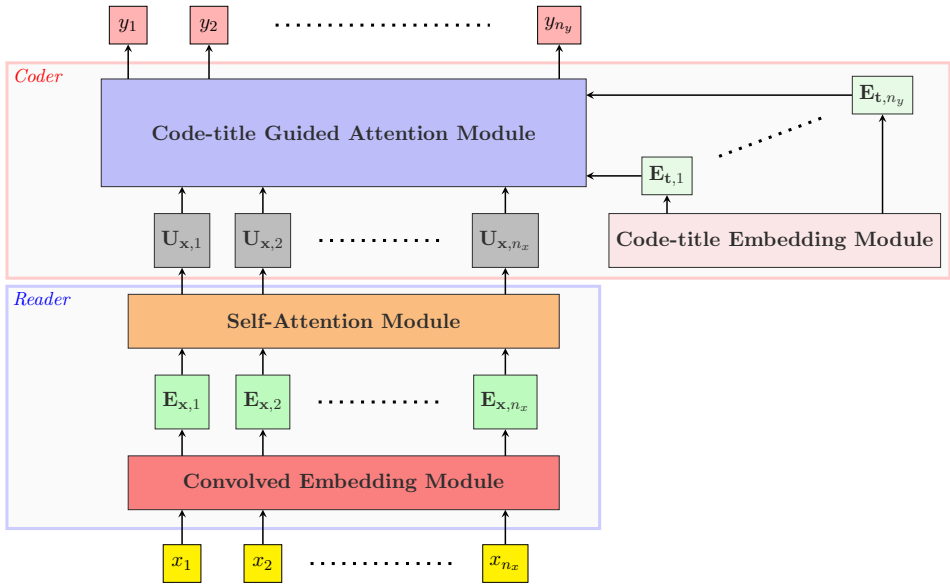
\begin{figure*}[ht]
    \centering
    \resizebox{1.0\textwidth}{!}{%
        \usetikzlibrary{shapes.geometric, shapes.multipart, arrows, calc, positioning}

\tikzset{
token_raw/.style= {draw, fill=yellow, rectangle, minimum width=0.7cm, minimum height=0.7cm, node distance=1cm},
token_emb/.style= {draw, fill=green!30, rectangle, minimum width=1.0cm, minimum height=1.0cm, node distance=1cm},
token_attn/.style= {draw, fill=black!30, rectangle, minimum width=1.0cm, minimum height=1.0cm, node distance=1cm},
token_out/.style= {draw, fill=red!30, rectangle, minimum width=0.7cm, minimum height=0.7cm, node distance=1cm},
token_desc/.style= {draw, fill=green!10, rectangle, minimum width=0.7cm, minimum height=0.7cm, node distance=1cm},
layer_emb/.style = {draw, fill=red!60, rectangle, minimum width=8.00cm, minimum height=1.0cm},
layer_attn/.style = {draw, fill=orange!60, rectangle, minimum width=8.00cm, minimum height=1.0cm},
layer_code/.style = {draw, fill=blue!30, rectangle, minimum width=9.00cm, minimum height=2.0cm},
layer_desc/.style = {draw, fill=red!10, rectangle, minimum width=3.00cm, minimum height=1.0cm},
box_read/.style = {draw=blue, ultra thick, fill=black!10, opacity=.2, rectangle, minimum width=11.0cm, minimum height=4.5cm},
box_code/.style = {draw=red, ultra thick, fill=black!10, opacity=.2, rectangle, minimum width=17.5cm, minimum height=4.0cm}
}

\tikzstyle{arrow} = [thick,->,>=stealth]

\begin{tikzpicture}[node distance=1.0cm]
    \tikzstyle{every node}=[font=\normalsize]
    
    \node (x1) [token_raw] {$x_{1}$};
    \node (x2) [token_raw, right of=x1, xshift=0.5cm] {\(x_{2}\)};
    \node (xm) [token_raw, right of=x2, xshift=3.5cm] {\(x_{n_{x}}\)};
    \node (emb)[layer_emb, above of=x1, xshift=3.0cm, yshift=0.5cm]{\textbf{Convolved Embedding Module}};
    
    \node (e1) [token_emb, above of=x1, yshift=2.0cm] {\(\textbf{E}_{\textbf{x},1}\)};
    \node (e2) [token_emb, above of=x2, yshift=2.0cm] {\(\textbf{E}_{\textbf{x},2}\)};
    \node (em) [token_emb, above of=xm, yshift=2.0cm] {\(\textbf{E}_{\textbf{x},n_{x}}\)};
    \node (selfattn) [layer_attn, above of=emb, xshift=0.0cm, yshift=2.0cm]{\textbf{Self-Attention Module}};
    
    \node (read)[box_read, above of=x1, xshift=2.25cm, yshift=1.9cm]{};
    \node[below right, text=blue] at (read.north west) {\small\textit{Reader}};

    \node (u1) [token_attn, above of=e1, yshift=2.0cm] {\(\textbf{U}_{\textbf{x},1}\)};
    \node (u2) [token_attn, above of=e2, yshift=2.0cm] {\(\textbf{U}_{\textbf{x},2}\)};
    \node (um) [token_attn, above of=em, yshift=2.0cm] {\(\textbf{U}_{\textbf{x},n_{x}}\)};
    \node (codeattn) [layer_code, above of=selfattn, xshift=0.0cm, yshift=2.5cm]{\textbf{Code-title Guided Attention Module}};
    
    \node (y1) [token_out, above of=u1, xshift=-1.0cm, yshift=3.00cm] {\(y_{1}\)};
    \node (y2) [token_out, right of=y1, xshift=.5cm] {\(y_{2}\)};
    \node (yn) [token_out, right of=y2, xshift=5.5cm] {\(y_{n_{y}}\)};
    
    \node (descemb) [layer_desc, right of=codeattn, xshift=7.0cm, yshift=-2.0cm]{\textbf{Code-title Embedding Module}};
    
    \node (t1) [token_desc, above of=descemb, xshift=-2.0cm, yshift=0.2cm] {\(\textbf{E}_{\textbf{t},1}\)};
    \node (tn) [token_desc, right of=t1, xshift=3.0cm, yshift=1.5cm] {\(\textbf{E}_{\textbf{t},n_{y}}\)};
    
    \node (code)[box_code, below of=y1, xshift=6.5cm, yshift=-1.7cm]{};
    \node[below right, text=red] at (code.north west) {\small\textit{Coder}};

    \draw [arrow] (x1) -- (0.0,1.0);
    \draw [arrow] (x2) -- (1.5,1.0);
    \draw [arrow] (xm) -- (6.0,1.0);

    \draw [arrow] (0.0,2.0) -- (e1);
    \draw [arrow] (1.5,2.0) -- (e2);
    \draw [arrow] (6.0,2.0) -- (em);

    \draw [arrow] (e1) -- (0.0,4.0);
    \draw [arrow] (e2) -- (1.5,4.0);
    \draw [arrow] (em) -- (6.0,4.0);

    \draw [arrow] (0.0,5.0) -- (u1);
    \draw [arrow] (1.5,5.0) -- (u2);
    \draw [arrow] (6.0,5.0) -- (um);
    
    \draw [arrow] (u1) -- (0.0,7.0);
    \draw [arrow] (u2) -- (1.5,7.0);
    \draw [arrow] (um) -- (6.0,7.0);

    \draw [arrow] (-1.0,9.0) -- (y1);
    \draw [arrow] (0.5,9.0) -- (y2);
    \draw [arrow] (7.0,9.0) -- (yn);
    
    \draw [arrow] (9.0,6.5) -- (t1);
    \draw [arrow] (t1) -- (7.5,7.2);
    \draw [arrow] (13.0,6.5) -- (tn);
    \draw [arrow] (tn) -- (7.5,8.7);

    \draw [loosely dotted, ultra thick, shorten <=15, shorten >=15] (x2) -- (xm);    
    \draw [loosely dotted, ultra thick, shorten <=15, shorten >=15] (e2) -- (em);    
    \draw [loosely dotted, ultra thick, shorten <=15, shorten >=15] (u2) -- (um);   
    \draw [loosely dotted, ultra thick, shorten <=25, shorten >=25] (y2) -- (yn);
    \draw [loosely dotted, ultra thick, shorten <=15, shorten >=15] (t1) -- (tn);    
    
\end{tikzpicture}
    }%
    \caption{Graphical visualization of Read, Attend, and Code (RAC) model architecture with the reader and coder modules, as described in Section \ref{sec:RAC Model}.}
    \label{fig:Arch}
\end{figure*}

\textbf{Baseline Results:} 
We calculated the similarity between the human coders we hired and the human coders that the BIDMC hired, available as a human reference in the MIMIC-III dataset, to estimate the human-coders inter-agreement as a human-coding baseline\footnote{We measured the human coding baseline by not comparing the agreement of the two certified coders we hired. Instead, we compared the coders we hired with the BIDMC coders. The BIDMC coders’ data is available as a reference label in the MIMIC-III dataset. Of course, neither is an absolute golden truth for inherent ambiguity in the human coding process. Alternatively, we could have two coders independently code, and the notes where they differed are coded independently by a third coder to collect golden reference. We were short on time to pursue this direction when preparing for submission.}. We also reported how much predicted annotations from our RAC model-based ML system agree with the same MIMIC-III reference to benchmark the ML system performance against the established human-level coding baseline. We summarized the measured human-coders inter-agreement rates in Table~\ref{table: human perf summary} in terms of the macro and micro Jaccard similarity, precision, and recall scores and compared them with our RAC-model based automated ML system. We found that estimated inter-coder agreement rates (i.e., human coding baseline) were not high as we initially thought and are far exceeded by the ML system's 3.9 times higher rate in Micro-Jaccard similarity. We are not here to claim that we reached super-human accuracy as there is room to improve the baseline; instead, this result highlights that our RAC-model based automated ML system is advanced enough to handle complex medical notes to help code more accurately.

% OK, so off the bat if our coders are finding things not on the list we know that they are disagreeing with the dataset. right now they are going with second best as instructed. I dont know the impact here, but do know one of them only got through 12 in a few hours so it cant be too high a volume.
% i might be missing something here but this sounds as if you are calculating the similarity between the human coders alpha health hired vs. between the human coders alpha health hired and the human coders from Beth Israel. not sure if it's possible to make it more clear that it's the latter

\section{RAC Model}
\label{sec:RAC Model}
In this section, we describe an end-to-end RAC model to solve the medical code prediction problem. Let $\mathcal{C}$ be the set of medical codes of size $n_{y}$, $\mathcal{C}_{T}$ be the set of concatenated long and short titles of all $n_{y}$ codes, $\mathbf{x}=(x_{1},...,x_{n_{x}})^{T}$ be the document with $n_{x}$ tokens, and $\mathbf{y}=(y_{1},...,y_{n_{y}})^{T}$ be a medical codes prediction vector. Our RAC model has two sub-modules: a reader and a coder. The reader is a self-attention module that takes a tokenized clinical note $\mathbf{x}$ as input, and the coder is a code-title guided attention module that predicts each medical code's likelihood vector $\mathbf{y}$. Figure~\ref{fig:Arch} visualizes RAC architecture with these components, and we provide more details in the following sections. 

\subsection{Read}
\label{ssec:Read}
\textbf{Convolved Embedding Module:} 
The \emph{Read} operation first converts each token into a token embedding of dimension $d$ by using an embedding layer followed by two one-dimensional CNN layers. Unlike prior works \citep{Mullenbach18, Xie19, Li20, Ji20}, we use CNN layers as an efficient preprocessing mechanism to capture an embedding for a group of nearby tokens (like n-grams), which is informative in the next self-attention module for expressing local dependencies in clinical notes. We let $\mathbf{E_{x}}\in\mathbb{R}^{n_{x}\times d}$ be the convolved embedding matrix corresponding to the $n_{x}$ tokens in the input document. Since the next self-attention module does not need to \say{know} the embeddings' orders, we do not need to concatenate with positional encoding, unlike the language modelings. 

We pre-train the Word2Vec Skip-gram model on all the notes in the training set using Gensim~\citep{Rehurek10}, with an embedding size of $d = 300$, a window size of 5, and a minimum word frequency count of 10 for 5 epochs. The pre-trained weights are then loaded into the embedding layer with a maximum input length of $n_{x}=4096$ tokens. We stack two CNN layers with $d$ filters, a kernel size of 10, and a tanh activation function on top of the embedding layer. We apply dropout to the module output with a rate of 0.1.

\textbf{Self-Attention Module:} 
In our model, the self-attention module (SAM) is a stack of four identical layers. Each layer is a stack of single-head self-attention and feed-forward layers interleaved with residual connections and layer normalization similar to~\citep{Vaswani17}. In particular, for the convolved embeddings $\mathbf{E_{x}}$, the dot product attention is computed as follows: 

\[\text{Attn}(\mathbf{E_{x}}) = \text{LN}\left(\mathbf{E_{x}}+\text{Softmax}\left[\frac{(\mathbf{E_{x}}\mathbf{W_{q}})(\mathbf{E_{x}}\mathbf{W_{k}})^{T}}{\sqrt{d}}\right](\mathbf{E_{x}}\mathbf{W_{v}})\right).\]
% \begin{multline*}
% \text{Attn}(\mathbf{E_{x}}) =\text{LN}\left(\mathbf{E_{x}}+ \right.\\
% \left.\text{Softmax}\left[\frac{(\mathbf{E_{x}}\mathbf{W_{q}})(\mathbf{E_{x}}\mathbf{W_{k}})^{T}}{\sqrt{d}}\right](\mathbf{E_{x}}\mathbf{W_{v}})\right).
% \end{multline*}

Here $\mathbf{W_{q}}, \mathbf{W_{k}}, \mathbf{W_{v}}\in\mathbb{R}^{d\times d}$ represent the projection matrices associated with the query, key, and value, respectively and \text{LN} represents the layer normalization. Subsequently, the layer's output takes the following form: 
% \[\text{FFN}(\mathbf{E_{x}})=\text{LN}\left(\text{Attn}(\mathbf{E_{x}})+\sigma(\text{Attn}(\mathbf{E_{x}})\mathbf{W_{1}})\mathbf{W_{2}}\right),\]
\[\text{LN}\left(\text{Attn}(\mathbf{E_{x}})+\sigma(\text{Attn}(\mathbf{E_{x}})\mathbf{W_{1}})\mathbf{W_{2}}\right),\]
where $\mathbf{W_{1}}\in\mathbb{R}^{d\times d_{ff}}$, $\mathbf{W_{2}}\in\mathbb{R}^{d_{ff}\times d}$, and $\sigma$ is a ReLU activation function. We use $d_{ff} = 1024$ and and apply dropout to each sub-layer's output before it is added to the sub-layer input and normalized with a rate of 0.1.

\subsection{Attend and Code}
\label{ssec:Attend and Code}
\textbf{Code-title Embedding Module:} 
We now illustrate the \emph{AttendAndCode} function to effectively handle the large code output space's extreme long-tail sparsity. The majority of medical codes are rare, \say{tail} codes. For example, 60\% of ICD codes appear less than 10 times in the MIMIC-III dataset. Moreover, there are roughly 16 codes on average in MIMIC-III discharge summaries, so 99.8\% are zeros in  $\mathbf{y}$. To handle this extreme sparsity of the large code output space, unlike prior sub-optimal attempts \citep{Mullenbach18,Wang18b,Xie19}, we utilize the code titles (i.e., output labels) information in terms of queries for the next attention module. Code titles corresponding with medical codes are defined by ICD coding systems. Such code titles include, for example, \say{Intestinal infection due to Clostridium difficile (008.45)} and \say{Sinoatrial node dysfunction (427.81)}.

% In the particular implementation, code titles corresponding with service codes may be defined according
% to a well-established set of service codes such as, for example, service codes defined by International Classification of Disease (ICD) coding systems. Such code titles may comprise text that uniquely identifies and/or is descriptive of services underlying associated service codes. Such code titles may include, for example, “gastric
% intubation” (43752, 91105); “interpretation of blood gases and interpretation of data stored in computers, such as ECGs, blood pressure, hematologic data” (99090); “interpretation of cardiac output” (93561-93562); “interpretation of chest X-rays” (71010-71020); “pulse oximetry” (94760-94762); “temporary transcutaneous pacing” (92953);
% “vascular access procedures” (36000, 36410, 36415, 36591, 36600); and “ventilator management” (94002-94004, 94660, 94662). In particular embodiments, service codes may be rare (aka “tail”). In particular implementations, application of code-title embedding 262 may enable a learning of semantic patterns and/or relationships
% between and/or among code titles that improves accuracy of code prediction.

We use the definition tables of the diagnoses and procedure codes, concatenate long and short titles together for all $n_{y}$ codes, and build $\mathcal{C}_{T}$ first. By tokenizing $\mathcal{C}_{T}$ with $n_{t}$ tokens, we have a title matrix $\mathbf{T}$ where $\mathbf{T}\in\mathbb{R}^{n_{y}\times n_{t}}$. From $\mathbf{T}$ input, the module extracts a code-title embedding of dimension $d$ by using an embedding layer followed by a single CNN layer and Global Max Pooling layer. We let $\mathbf{E_{t}}\in\mathbb{R}^{n_{y}\times d}$ be the extracted code-title embedding matrix. In the model, each concatenated code title is padded to $n_{t}=36$ tokens, the same pre-trained Word2Vec Skip-gram model weights that the reader used are loaded to initialize the embedding layer, and a single CNN layer with $d$ filters, kernel size 10, and tanh activation function are used.

\textbf{Code-title Guided Attention Module:} 
This function computes code-level attention over the reader output to attend to different parts for each code. We explicitly use $\mathbf{E_{t}}$ as a query matrix to guide where to attend from the reader output. Specifically, the approach leads to the following attention mechanism:

\[\mathbf{V_{x}}=\text{Softmax}\left(\frac{\mathbf{E_{t}}\mathbf{U^{T}_{x}}}{\sqrt{d}}\right)\mathbf{U_{x}},\]
where $\mathbf{U_{x}}\triangleq\text{SAM}(\mathbf{E_{x}})$ and $\mathbf{V_{x}}\in\mathbb{R}^{n_{y}\times d}$. 

Fundamental improvement from the per-label attention initially introduced in~\citep{Mullenbach18} is that $\mathbf{E_{t}}$ is no longer randomly initialized. It is straightforward to show that queries close in Euclidean space have similar attention scores, and this property is very desirable for learning inter-relations between less frequent codes and the text effectively. Using $\mathbf{E_{t}}$ as a query in computing attention scores has actually resulted in large performance gains, as clearly seen by comparing the first and second row in the RAC Models section of Table~\ref{table: machine perf summary}. This reveals that using learned semantic patterns of code titles in query leads to improving the RAC model's quality, particularly for the \say{tail} codes.

With attended $\mathbf{V_{x}}$, finally, the module produces a code likelihood vector $\mathbf{y}$ as 
\[\mathbf{y}=\sigma(\mathbf{V_{x}}\mathbf{W_{3}}),\]
where $\mathbf{W_{3}}\in\mathbb{R}^{d\times 1}$ and $\sigma$ is the sigmoid function.

\subsection{Learn the Entire Model}
\label{ssec:Learn the Entire Model}
\textbf{Sentence Permutation:} Unlike computer vision tasks, data augmentation should be done carefully in the NLP application due to the text's structure. By taking into account the problem's nature of permutation equivariance, we rely on a simple sentence permutation method. Provided that notes in the training set contain multiple sentences, they are shuffled in a random order to generate a new train sample with the same label. For our model training, we use the 3-fold augmentation to increase the training set size three times and conclude that it is beneficial, as seen by comparing the second and third rows in the RAC Models section of Table~\ref{table: machine perf summary}. One hindsight here is that the second row shows the RAC model's clear wins over the best prior baselines across the board without providing the augmented data. In other words, this indicates that the performance improvements of the RAC model are not simply the effects of the sentence permutation. 

\textbf{Stochastic Weighted Averaging (SWA):} 
With the augmented train data ready, the entire RAC model is trained on the medical codes prediction task to maximize the log-likelihood of $n_{y}$ binary classifiers. Rather than using traditional ensembling techniques to combine multiple models to make an averaged final prediction, we apply the SWA approach. Using SWA, we store the running average of model weights during training, so prediction is faster than conventional ensemble methods. In our training, SWA is averaged every 5 epochs from the first epoch, which provides positive gains, as noted by comparing the last two rows in Table~\ref{table: machine perf summary}. 

\begin{table*}[ht]\small
    \caption{Medical codes prediction results (in \%) by ML systems on the MIMIC-III-full-label testing set. The bold value shows the best (and highest) SOTA value for each column metric, and underlined numbers indicate the previous SOTA result. The proposed RAC model outperforms all the previously reported SOTA models and achieves new SOTA milestones across all the evaluation metrics. Note that the RAC model has shown an 18.7\% relative performance increase in Macro-F1 over the prior best SOTA model. Results for the baseline models are taken from~\citep{Mullenbach18}.}
    \label{table: machine perf summary}
    \centering
    \begin{tabular}[t]{l|l|l|l}
    \toprule[\heavyrulewidth]
    \textbf{Model} & \textbf{AUC} & \textbf{F1} & \textbf{Precision@n} \\
    & Macro Micro & Macro Micro & 5 \ \ \ \ \ \ \ 8 \ \ \ \ \ \ \ 15 \\
    \midrule
    \textbf{Baseline Models}\\
    Logistic Regression & 56.1 \ \ \ 93.7 & \ 1.1 \ \ \ \ \ 27.2 & \ \ \ \ \ \ \ \ \ 54.2 \ \ \ 41.1\\ 
    SVM & \ \ \ \ \ \ \ \ \ \ \ \ \ \ \ \ \ & \ \ \ \ \ \ \ \ \ \ \ 44.1 & \ \ \ \ \ \ \ \ \ \ \ \ \ \ \ \ \ \ \ \ \ \ \ \ \ \\ 
    CNN & 80.6 \ \ \ 96.9 & \ 4.2 \ \ \ \ \ 41.9 & \ \ \ \ \ \ \ \ \ 58.1 \ \ \ 44.3 \\ 
    Bi-GRU & 82.2 \ \ \ 97.1 & \ 3.8 \ \ \ \ \ 41.7 & \ \ \ \ \ \ \ \ \ 58.5 \ \ \ 44.5 \\ 
    \midrule
    \textbf{CNN-based Models}\\
    CAML \citep{Mullenbach18} & 89.5 \ \ \ 98.6 & \ 8.8 \ \ \ \ \ 53.9 & \ \ \ \ \ \ \ \ \ 70.9 \ \ \ 56.1 \\ 
    DR-CAML \citep{Mullenbach18} & 89.7 \ \ \ 98.5 & \ 8.6 \ \ \ \ \ 52.9 & \ \ \ \ \ \ \ \ \ 69.0 \ \ \ 54.8 \\ 
    MSATT-KG \citep{Xie19} & 91.0 \ \ \ \textbf{99.2} & \ 9.0 \ \ \ \ \ 55.3 & \ \ \ \ \ \ \ \ \ 72.8 \ \ \ 58.1 \\ 
    MultiResCNN \citep{Li20} & 91.0 \ \ \ 98.6 & \ 8.5 \ \ \ \ \ 55.2 & \ \ \ \ \ \ \ \ \ 73.4 \ \ \ 58.4 \\ 
    \midrule
    \textbf{LSTM-based Models}\\
    LAAT \citep{Vu20} & 91.9 \ \ \ \underline{98.8} & \ 9.9 \ \ \ \ \ \underline{57.5} & \underline{81.3} \ \ \ \underline{73.8} \ \ \ \underline{59.1} \\ 
    JointLAAT \citep{Vu20} & \underline{92.1} \ \ \ \underline{98.8} & \underline{10.7} \ \ \ \ \underline{57.5} & 80.6 \ \ \ 73.5 \ \ \ 59.0 \\ 
    \midrule
    \textbf{RAC Models (ours)}\\ 
    RAC - SWA - SentPerm - CodeTitles & 92.4 \ \ \ 98.9 & 11.0 \ \ \ \ 57.4 & 81.5 \ \ \ 74.2 \ \ \ 58.8 \\ 
    RAC - SWA - SentPerm & \textbf{94.9} \ \ \textbf{99.2} & 11.4 \ \ \ \ 58.0 & 82.3 \ \ \ 74.9 \ \ \ 59.5 \\ 
    RAC - SWA & 94.8 \ \ \ \textbf{99.2} & 12.6 \ \ \ \ 58.2 & 82.6 \ \ \ 74.9 \ \ \ 59.8 \\
    RAC & 94.8 \ \ \ \textbf{99.2} & \textbf{12.7} \ \ \ \textbf{58.6} & \textbf{82.9} \ \ \textbf{75.4} \ \ \textbf{60.1} \\ 
    \bottomrule[\heavyrulewidth]
    \end{tabular}
\end{table*}

\section{Codes Prediction by Machines} 
\label{sec:Experimental Results}
\textbf{MIMIC-III Dataset:} The MIMIC-III Dataset (MIMIC v1.4) is a freely accessible medical database containing de-identified medical data of over 40,000 patients staying in the Beth Israel Deaconess Medical Center between 2001 and 2012. For this study, we extract the discharge summaries and the corresponding ICD-9 codes\footnote{There are two reasons that the ICD-9 codes are used in the experimental studies. First, the publicly accessible MIMIC-III dataset used for the study was collected between 2001 and 2012 before the ICD-10 adoption. Second, the MIMIC-III dataset has been used as a standard benchmark in the prior studies, making meaningful head-to-head comparisons with our work. Additionally, we believe the proposed RAC model is not limited to the ICD-9 system. It is anticipated to work for a more complex and sparse environment (like ICD-10 and forthcoming ICD-11 codesets) with the help of employing code-title guided attention. The advantage of employing the RAC model (vs. prior arts) will be even more pronounced.}. We perform the same data processing, and data split as stated in \citep{Mullenbach18} for a direct comparison with prior works. This processing results in  47,724 samples for training, 1,632 and 3,373 samples for validation and testing. One can refer to Table 2 in \citep{Mullenbach18} for more dataset statistics. 
% Even though slightly different preprocessing might help the performance, all punctuation marks, special characters, and tokens with no alphabetic characters are removed.

\textbf{Training Details:} 
For this study, we use an Adam optimizer with a learning rate of 8e-5 and pick a batch size of 16 on 4 Nvidia T4 GPUs. All model training runs are carried with the early stopping after the Precision@8 on the validation set does not improve for 3 epochs. The RAC model typically converges within 10 epochs. The model at the epoch of the highest Precision@8 is evaluated on the testing set. Precision@n corresponds to the number of ground truth labels among the n top-scored outputs, and we made this choice because Precision@8 is informative in a production scenario of providing a fixed number of codes to review.

\textbf{Performance Results:}
We predict a total of 8,921 unique ICD-9 codes (composed of 6,918 diagnosis and 2,003 procedure codes) available in the MIMIC-III-full-label dataset. A smaller subset of code prediction tasks (e.g., TOP-10 or TOP-50 frequent codes) is not considered because it is less challenging and has limited value in our production scenario. We summarized all the previously reported SOTAs values in Table~\ref{table: machine perf summary} in terms of the macro and micro average AUC, F1, and Precision@n scores to compare with prior works. Note that the RAC model's number can be further optimized for each metric. Table~\ref{table: machine perf summary} shows the RAC models' superior performance over the past SOTA models across the board. What stands out is that the RAC model show very sharp improvements, 18.7\% at Macro-F1 and 3.0\% at Macro-AUC. These advancements imply that the RAC model learns a better mapping from the clinical text input to the infrequent codes since macro-metrics places more weight on the uncommon codes. Comparing the last four rows in Table~\ref{table: machine perf summary} highlights each component's contributions to the overall performance gains in the RAC model. 

\section{Discussion} 
In this paper, we present for the first time a human-coding baseline for medical code prediction on the subsampled MIMIC-III-full-label inpatient clinical notes testing set task. We have developed an attention-based RAC model that sets the new SOTA records, and the resulting RAC model outperforms the human-coding baseline to a great extent on the same task. The performance improvements can be attributed to effectively learning the common embedding space between the clinical note and medical codes by utilizing attention mechanisms that efficiently address the severe long-tail sparsity issues. This achievement is one step forward to the bigger vision of a fully autonomous ML coding system that autonomously codes the medical charts without input from human coders. 
% integrating an automated ML system with human coder capabilities for medical code prediction. 

\textbf{Limitations:} 
The current evaluation has the disadvantage of taking only discharge summaries out of the entire inpatient medical chart. We could turn to advanced models to handle more complex inpatient charts containing various service records. We have not discussed how much the RAC model's accurate prediction performance positively impacts the coding professionals and the healthcare delivery organizations. Conceivably, the computed attention scores can be visualized to inform the notes' relevant portions to understand the predictions. We intend to evaluate how much this visualized insight along with the presentation of likely medical codes would help coding teams not waste their time and resources as a future study. 

\textbf{Ethics and Broader Impact:} 
An automated ML system for medical code prediction, first of all, intends to streamline the medical coding workflow, reduce human coders' backlog by increasing productivity, and help human coders navigate through extended and complex charts quickly while reducing coding errors~\citep{Crawford13}. Suppose relevant diagnoses and procedures are annotated with the appropriate codes, human coders can review the record much more quickly and validate the correct codes. Secondly, the automated system intends to reduce the administrative burden on providers, who could instead focus on delivering care rather than learning the nuances of coding. It also helps maximize revenue by capitalizing on the level of specificity noted in the documentation. This will impact many corners of the medical claims billing process by improving coding accuracy and consistency and reducing denials and compliance risk. Moreover, better-automated software can further enhance clinical documentation, making the overall picture of its quality better, eventually redirecting the wasted healthcare costs to more meaningful purposes~\citep{Shrank19}.

% ACKNOWLEDGEMENTS ONLY GO IN THE CAMERA-READY, NOT THE SUBMISSION
\acks{During this work, the discussions and actual human coding baseline projects held with Amy Raymond and Grant Messick proved stimulating and helpful. The author also would like to thank Jesse Swidler, Peng Su, and Shan Huang for considerable infrastructure support. The suggestions of Heidi Lim to improve the figures and website visualizations are also gratefully acknowledged.}

% ACKNOWLEDGEMENTS ONLY GO IN THE CAMERA-READY, NOT THE SUBMISSION
% \acks{Many thanks to all collaborators and funders!}

\bibliography{custom}

% \appendix
% \section*{Appendix A.}

% Some more details about those methods, so we can actually reproduce
% them.  After the blind review period, you could link to a repository
% for the code also.  \emph{MLHC values both rigorous evaluation as well
%   as reproduciblity.}

\end{document}